\documentclass[OTHER]{cta-author}
\usepackage{hyperref}
\usepackage{multirow}
\usepackage{tabularray}
\usepackage{caption}
\usepackage{pgfplots} 
\usepackage{booktabs}
\usepackage{tabu}
\usepackage{subcaption}
\pgfplotsset{compat=newest} 

\usepackage[acronym,nomain,nonumberlist]{glossaries}
\makeglossaries

\newacronym{norppa}{NORPPA}{NOvel Ringed seal Re-identification by Pelage Pattern Aggregation}
\newacronym{gva}{GVA}{Geometric Verification Algorithm}
\newacronym{est}{$\text{E}_\Delta=f(d, n, \omega)$}{mathematical abbreviation for GVA}
\newacronym{sealid}{SealID}{Saimaa Ringed Seal Re-Identification Dataset}
\newacronym{ransac}{RANSAC}{Random Sample Consensus}
\newacronym{fv}{FV}{Fisher Vector}
\newacronym{pca}{PCA}{Principal Component Analysis}
\newacronym{gmm}{GMM}{Gaussian Mixture Model}
\newacronym{2d}{2D}{two-dimensional}
\newacronym{3d}{3D}{three-dimensional}
\newacronym{svd}{SVD}{Singular Value Decomposition}
\newacronym{cab}{CAB}{Cross-Attention Block}
\newacronym{owt}{OWT}{Oriented Watershed Transform}
\newacronym{umc}{UMC}{Ultrametric Contour Map}
\newacronym{lnbnn}{LNBNN}{Local Naive Bayes Nearest Neighbor}
\newacronym{sift}{SIFT}{Scale Invariant Feature Transform}
\newacronym{sfta}{SFTA}{Segmentation-based Fractal Texture Analysis}

\begin{document}

%%%%%%%%%%%%%% title and the authors %%%%%%%%%%

\supertitle{Brief Paper}

\title{Combining feature aggregation and geometric similarity for re-identification of patterned animals}

\author{
	\au{Veikka Immonen$^1$}
	\au{Ekaterina Nepovinnykh$^{1,*}$}
    \au{Tuomas Eerola$^1$}
	\au{Charles V. Stewart$^2$}
    \au{Heikki K\"{a}lvi\"{a}inen$^1$}
}

\address{
	\add{1}{Computer Vision and Pattern Recognition Laboratory, Department of Computational Engineering, School of Engineering Sciences, Lappeenranta-Lahti University of Technology LUT, 
    %P.O. Box 20, 
    FI-53851 Lappeenranta, Finland}
	\add{2}{Department of Computer Science, Rensselaer Polytechnic Institute, Troy, NY 12180, USA}
	\email{ekaterina.nepovinnykh@lut.fi}
}

%%%%%%%%%% end title and the authors %%%%%%%%%%

%%%%%%%%%%%%%% abstract %%%%%%%%%%

\begin{abstract}
%This should be informative and suitable for direct inclusion in abstracting services as a self-contained article. It should not exceed 200 words. It should summarise the general scope and also state the main results obtained, methods used, the value of the work and the conclusions drawn. No figure numbers, table numbers, references or displayed mathematical expressions should be included.
Image-based re-identification of animal individuals allows gathering of information such as migration patterns of the animals over time. This, together with large image volumes collected using camera traps and crowdsourcing, opens novel possibilities to study animal populations. For many species, the re-identification can be done by analyzing the permanent fur, feather, or skin patterns that are unique to each individual. In this paper, we address the re-identification by combining two types of pattern similarity metrics: 1) pattern appearance similarity obtained by pattern feature aggregation and 2) geometric pattern similarity obtained by analyzing the geometric consistency of pattern similarities. The proposed combination allows to efficiently utilize both the local and global pattern features, providing a general re-identification approach that can be applied to a wide variety of different pattern types. In the experimental part of the work, we demonstrate that the method achieves promising re-identification accuracies for Saimaa ringed seals and whale sharks.
\end{abstract}

\maketitle

%%%%%%%%%% end abstract %%%%%%%%%%

%%%%%%%%%%%%%% main matter %%%%%%%%%%

\section{Introduction}
\label{sec:intro}

Automatic camera traps allow the collection of large volumes of wildlife images in a non-invasive way. To fully utilize this data in the research on animal populations, the analysis of the images needs to be automated. The essential image analysis problem to be solved is the re-identification of individual animals as it allows us to obtain, e.g., information about the behavior and migration patterns, as well as, estimate the population size through capture and recapture analysis.
The re-identification can be done by utilizing permanent visual traits such as fur pattern, scarring, or fin shape.

Re-identification has already been applied to study various animal species. For example, for Saimaa ringed seals (\emph{Pusa hispida saimensis}), an endangered species native to Lake Saimaa in Finland,  image-based re-identification has been applied for conservation purposes to study animal migration and behavior~\citep{koivuniemi2016photo, koivuniemi2019mark, kunnasranta2021sealed}. Recently, many automated Saimaa ringed seal re-identification methods utilizing the permanent ring pattern have been developed~\cite{nepovinnykh2020siamese,seal_cnn,nepovinnykh2023re}.
% Whale shark
Similarly, population monitoring efforts of whale sharks (\emph{rhincodon typus}) have been tackled using re-identification and capture and recapture models~\cite{rohner2022need}, and automatic methods for re-identification have been proposed~\cite{arzoumanian2005astronomical,kholiavchenko2022comprehensive}. These methods utilize the spot patterns that are unique to each individual. Many of the existing re-identification methods are specific to certain animal species or types of patterns, limiting the breadth of this their usability. General methods that can be applied to a wide variety of different animal species would be preferable as they could be easily adapted for various animal population studies.

\begin{figure}[!t]
\centering{\includegraphics[width=\linewidth]{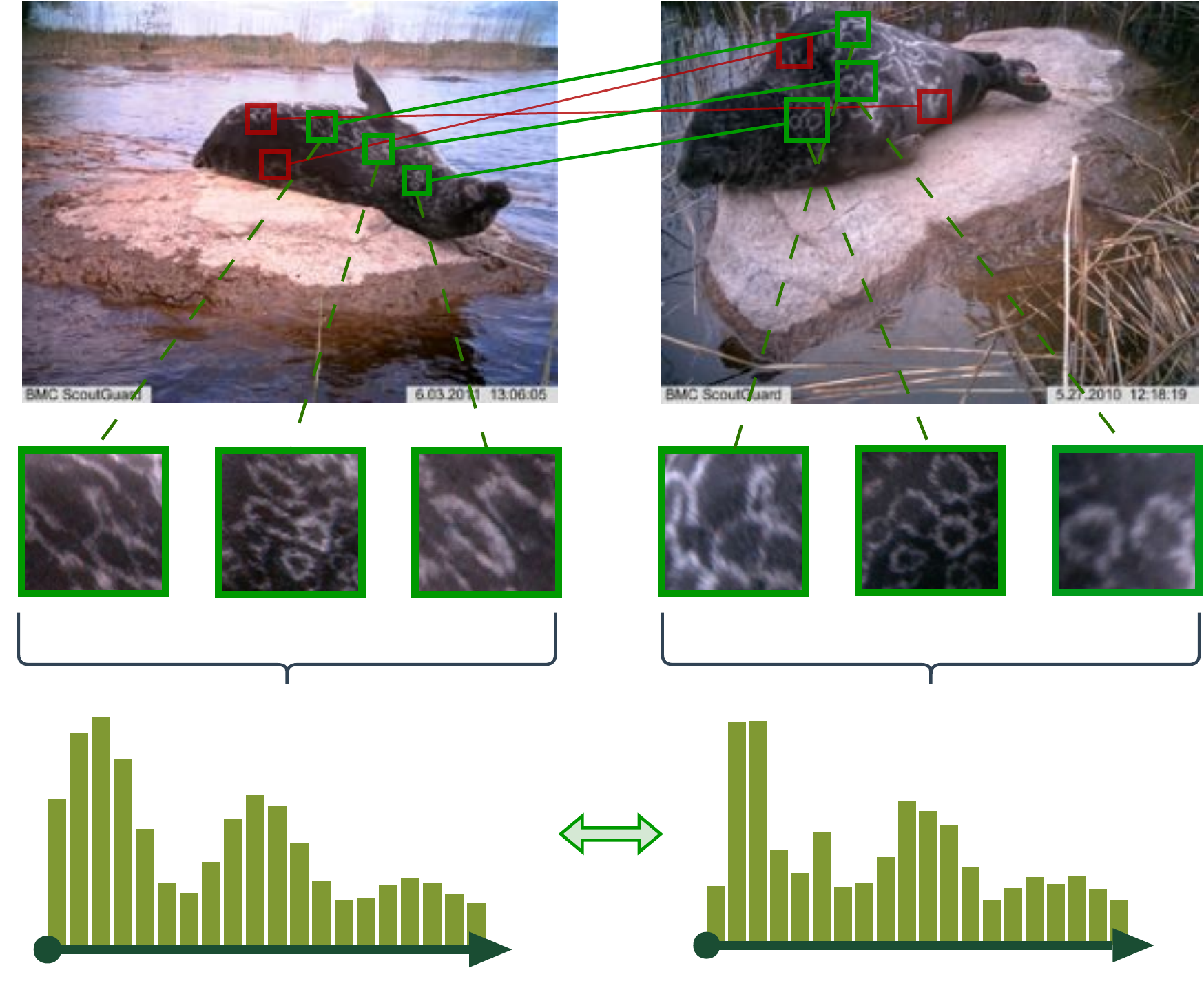}}
\caption{The proposed method combines aggregated pattern features and geometric consistency of the point correspondences. \label{fig:intro}}
\end{figure}

In this paper, species-agnostic re-identification is addressed by proposing a method that combines both local and global similarities of the fur pattern (see Fig.~\ref{fig:intro}). Local similarities are obtained by aggregating CNN-based local pattern features over a processed image using Fisher vectors. This results in compact vector presentations of the pattern appearance for the query images enabling efficient similarity comparison to the corresponding Fisher vectors computed for images in the database of known individuals. Fisher vector based aggregation does not, however, take into account the geometric consistency of the pattern similarity. Therefore, the similarity of the aggregated local pattern appearance is further complemented with a geometric similarity providing the global context for the pattern analysis. The geometric similarity is obtained by searching the local pattern feature correspondences between the query and database images and analyzing how consistent they are with the geometric transformation estimated using RANSAC. 

In the experimental part of the work, we demonstrate the accuracy of the proposed method on two species with very different types of patterns: Saimaa ringed seals and whale sharks. We show that the proposed method outperforms two earlier species-agnostic methods: HotSpotter~\cite{crall2013hotspotter} and NORPPA~\cite{seal_cnn}. Furthermore, we show that the combination of aggregated pattern feature-based similarity and geometric similarity provides higher re-identification accuracy than either of the similarity metrics alone. One notable benefit of the proposed pattern feature extraction method is that it does not require species-specific training but the same pre-trained keypoint detection and feature descriptors can be used for both species, making it a promising approach for generic animal individual re-identification.

\section{Related work}
\label{sec:related}

Various animal species can be re-identified by different types of visually unique biological traits such as fur pattern, face, or fin shape. Algorithmically, the methods can be divided into classification and metric-based approaches~\citep{vidal2021perspectives}. The classification-based approaches assume that the database of known individuals is known and finite. The metric-based methods, on the other hand, aim to learn a similarity metric between the input images. The re-identification is then performed by matching based on the similarity which means that metric-based approaches are not limited by the initial database and can be applied to new individuals without retraining. 
A variety of methods for animal re-identification exist. They have been successfully applied, for example, to Amur tigers (stripe pattern)~\citep{Liu_2019_ICCV}, cattle (muzzle shape)~\citep{kumar2018deep}, giraffes (spot pattern)~\citep{miele2021revisiting}, humpback whales (fluke shape)~\citep{weideman2020extracting} and primates (face)~\citep{deb2018face}.

% Saimaa ringed seals
A number of methods for the re-identification of Saimaa ringed seals have been proposed~\citep{zhelezniakov2015segmentation,chehrsimin2018automatic,nepovinnykh2018identification,nepovinnykh2020siamese,chelak2021eden,seal_cnn,nepovinnykh2023re}.
Saimaa ringed seal is especially challenging species for re-identification due to the following matters: (i) the large variation in possible poses which is further exacerbated by the deformable nature of the seals, (ii) the non-uniform pelage patterns, limiting the size of the regions that can be used for the re-identification task, (iii) the low contrast between the ring pattern and the rest of the pelage, and (iv) the extreme dataset bias. These challenges have been addressed by proposing various approaches to preprocess the images and to encode the pattern features~\cite{zhelezniakov2015segmentation,nepovinnykh2020siamese,chelak2021eden,seal_cnn}. The most successful methods employ the pattern extraction step~\cite{nepovinnykh2020siamese,seal_cnn} to construct a binary representation of the pelage pattern and metric learning-based pattern encoding.

% whale shark re-identification
Individual whale sharks can be identified based on the spot pattern on their skin. In~\cite{arzoumanian2005astronomical}, a blob detection was applied to find the individual spots, and a pattern-matching algorithm~\cite{groth1986pattern} originally developed for astronomical images (star patterns) were used to compare the patterns. 
In~\cite{kholiavchenko2022comprehensive}, a U-Net-based model was utilized for spot detection and a metric learning-based approach generated pattern embeddings for the re-identification of individuals.

% species-agnostic re-identification
While the majority of the existing methods are specific to species, there has been also efforts towards species-agnostic re-identification methods that can be applied to a wide variety of different type visual traits. HotSpotter~\citep{crall2013hotspotter} is a SIFT-based algorithm that uses viewpoint invariant descriptors and a scoring mechanism which emphasizes the most distinctive key points, called “hot spots,” on an animal pattern. PIE~\citep{moskvyak2019robust} is a deep learning-based method that receives shape embedding and pose embedding separately and normalizes the shape to match the individual regardless of the specific pose. In~\cite{schneider2022similarity}, various similarity learning architectures are compared on chimpanzees, humpback whales, fruit flies, and Siberian tigers.

\section{Proposed method}
\label{sec:proposed}

\begin{figure*}[!t]
\centering{\includegraphics[width=0.8\linewidth]{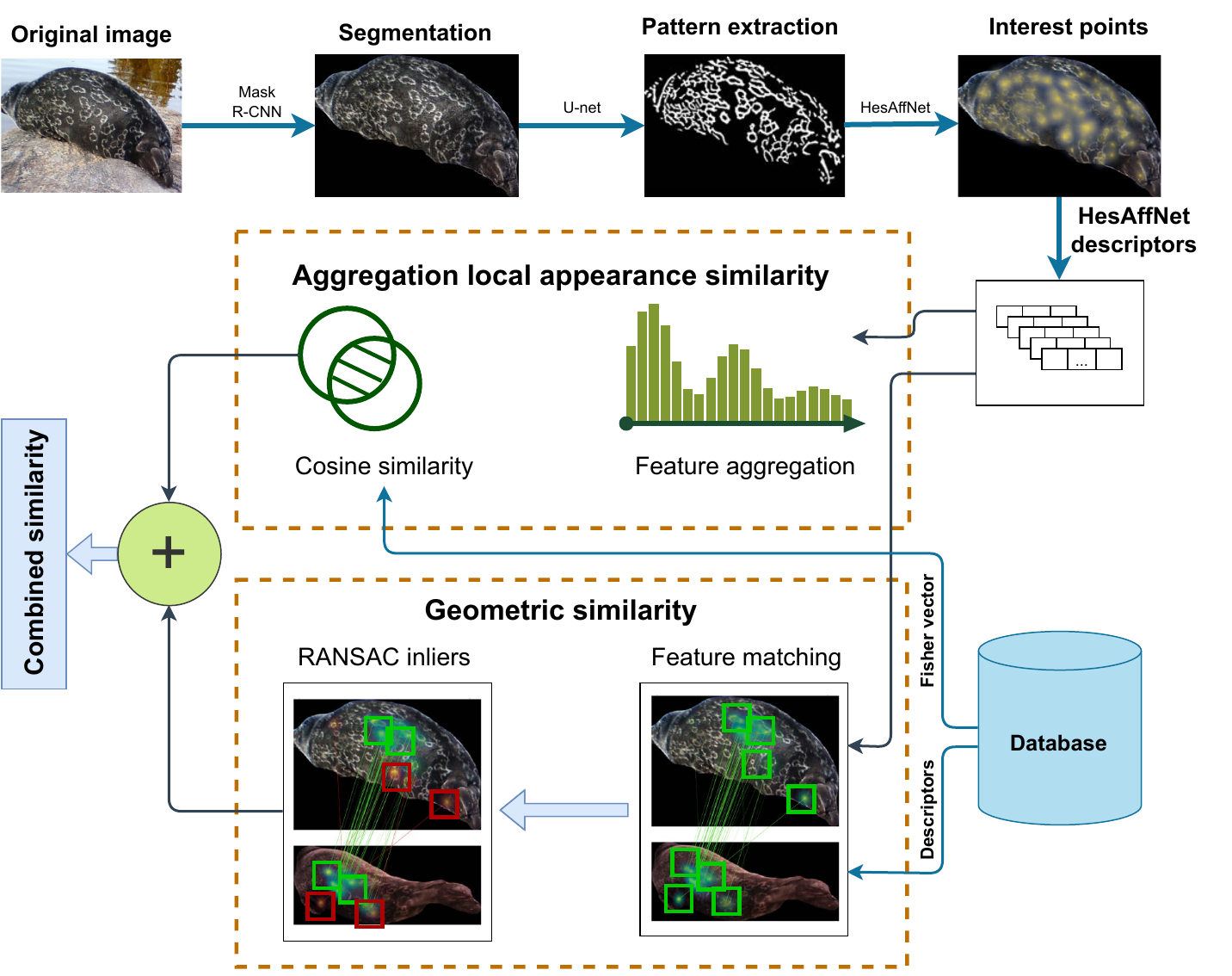}}
\caption{The pipeline of the proposed method. \label{fig:method}}
\end{figure*}

The proposed method builds on the NORPPA method~\cite{seal_cnn} developed for Saimaa ringed seals. We generalize the method to other patterned animal species by extending the pattern similarity method to address both the similarities in local appearance and the global geometric consistency of the patterns. The method starts with the segmentation of the animal from the background, after which the pattern is extracted for further analysis. Regions of interest (patches) are detected from the pattern images and the pattern image patches are embedded. To measure the similarity of the local pattern appearance, embeddings are aggregated to a single Fisher vector of a fixed length. These vectors can be used to quantify the similarity of two patterns using the cosine distance.
Since Fisher vectors do not take into account the global spatial structure of the pattern further geometric similarity is assessed by analyzing the consistency matched point correspondences to the homography found using RANSAC. Finally, the two similarity metrics are combined and the most similar pattern is searched from the database of known individuals. The whole pipeline is visualized in Fig.~\ref{fig:method}.

\subsection{Preprocessing and segmentation}
% tonemapping + segmentation (+ pattern extraction?)

Preprocessing of the data consists of two steps: tone mapping and segmentation. Images collected using camera traps and crowdsourcing contain a large variation in the quality due to challenging illumination, weather conditions, and suboptimal quality of cameras. To address this, we utilize tone mapping to balance the contrast levels, making it easier to segment the animal and extract the pelage pattern. For tone mapping, we utilize the method by Mantiuk et al.~\cite{mantiuk_perceptual_2006}. 
Various animals exhibit strong site-fidelity, meaning that they tend to stay in the same regions. This together with the static camera traps can lead to dataset bias issues as individuals are often captured with the same background increasing the risk that a supervised recognition algorithm learns to identify the background instead of the animal.
This is why segmentation of the animal is recommended. For segmentation, we utilize Mask R-CNN~\cite{he2017mask}.

\subsection{Aggregated local pattern appearance}
% HesAffNet + HardNet features
% feature aggregation
% (cosine) similarity of aggregated local pattern appearance

To analyze the local pattern appearances, local features are extracted and aggregated into comparable feature embeddings. First, the pattern is extracted from the segmented images with a pre-trained U-Net-based encoder-decoder model to emphasize the pattern and discard irrelevant information. Next, affine covariant regions are found and extracted from the pattern image by CNN-based HesAffNet~\cite{mishkin2018repeatability}. HesAffNet extracts local regions from images and transforms them according to the estimated local affine transformation, generating affine-invariant patches of fur patterns. 
% HesAffNet is trained by using the HardNegC loss function~\cite{mishkin2018repeatability} similar as the triplet loss. The HardNegC loss is defined as~\cite{mishkin2018repeatability}
% %
% \begin{equation}
%     L = \frac{1}{n} \sum_{i=1,n}{\max(0, 1 + d(s_i, \dot s_i) - d(s_i, N))}, \quad\frac{\partial L}{\partial N} := 0,
%     \label{eq:hardnegc}
% \end{equation}
% %
% where $d(s_i, \dot s_i)$ is the distance between matching patches and $N$ is the hardest negative sample in the training batch, making $d(s_i, N)$ the distance to that sample.

The extracted patches are embedded into vectors of size 1$\times$128 by using HardNet~\cite{mishchuk2017working}. 
HardNet is trained to correctly match corresponding descriptors while avoiding false positives from similar descriptors by using the triplet margin loss. 
% which is defined as~\cite{mishchuk2017working}
% %
% \begin{equation}
%     L = \frac{1}{n} \sum_{i=1,n}{\max(0, 1 + d(a_i, p_i) - \min(d(a_i, p_{j_{\min}}), d(a_{k_{\min}}, p_i)))},
%     \label{eq:hardnettriplet}
% \end{equation}
% %
% where $d$ is the chosen distance function for measuring the distance between descriptors, $a_i$ is the reference descriptor from a group of descriptors $A$, $p_i$ is from another group of descriptors $P$ and is a positive match to $a_i$, $p_{j_{min}}$ is the closest negative match to $a_i$ from $P$ and $a_{k_{min}}$ is the closest negative match to $p_i$ from $A$. Choosing the minimum distance between the positive and negative matches in the loss function effectively always picks the most difficult sample into the triplet, improving the model's ability to avoid false positives. 
After the HardNet embedding, PCA is applied to the features to decorrelate them and to reduce dimensionality.

Feature embeddings are aggregated using Fisher vectors~\citep{perronnin2007fisher, perronnin2010large, perronnin2010improving}. A visual vocabulary is constructed by applying Gaussian Mixture Model (GMM) to the features from the database. Then, Fisher vectors are created for each image by computing the partial derivatives of the log-likelihood function with respect to the GMM parameters and concatenating them. Kernel PCA~\citep{scholkopf_nonlinear_1998} is applied to further reduce the dimensionality of the resulting image descriptors which helps to reduce the storage requirements for the database, as well as speed up the database search for the re-identification.

Finally, (dis)similarities between the Fisher vector for the query image and the database images are calculated using the cosine distance:
\begin{equation}
d_L = 1-\frac{\Phi_q \cdot \Phi_{db}}{\lvert \lvert \Phi_q \rvert \rvert _2 \lvert \lvert \Phi_{db} \rvert \rvert _2}
\end{equation}
where $\Phi_q$ is the Fisher vector for query image and $\Phi_{db}$ the Fisher vector for a database image. This distance quantifies the dissimilarity of the aggregated local pattern appearances between the images. For more detailed description of the pattern feature aggregation, see~\cite{seal_cnn}. 

\subsection{Geometric similarity of patterns} 
% Veikka's geometric verification method

The aggregated local pattern appearance does not take into account the global spatial structure of the pattern. To further incorporate this information to the pattern matching, the geometric consistency of the local similarities are analyzed. This is done using a similar method as the spatial reranking step of the HotSpotter algorithm~\citep{crall2013hotspotter} and the object retrieval method proposed in~\cite{philbin2007object}. The HardNet embeddings of the HesAffNet interest points are matched to find the patch correspondences between query and database images. The matching is done by computing cosine distances between the embeddings of individual patch pairs.

The coordinates of the patch correspondences are then normalized to have the zero mean and the maximum distance of 1 to the origin. Outliers (and inliers) are detected by estimating the projective homography between the query image and database image using RANSAC. The assumption is that if the patterns do not match, the inconsistency in the global arrangements of patch correspondences causes a low amount of inliers. Therefore, the amount ($n$) and relative proportion ($\omega$) of inliers are good metrics for geometric similarity of patterns. It should be noted that due to the large pose variation of animals, it is recommended to have a high inlier threshold to ensure successful outlier detection in the case of matching patterns.

\subsection{Combined similarity}
% the combination rule

The final re-identification of the animal individual in the query image is performed by searching the most similar pattern from the database of known individuals. To compute the dissimilarity (distance) a novel combination of the dissimilarity of aggregated local pattern appearance and geometric dissimilarity of patterns is used. We propose two combination rules:
\begin{equation}
    d_C = d_L (1 - \omega)^a
\end{equation}
and
\begin{equation}
    d_C = d_L ^ n.
\end{equation}
%where $d_L$ is the cosine distance between the Fisher vectors (aggregated local pattern appearance), $\omega$ is the ratio of inliers to all point correspondences, and $n$ is the number of inliers. 
In the first combination rule, the geometric similarity, defined as a ratio of inliers to all image points $\omega$, has a polynomial influence on the cosine distance $d$ between the Fisher vectors (aggregated local pattern appearance). The greater the value of $a$ ($a \geq 0$) is, the more influence geometric consistency has in the final results. In the second combination rule, the geometric consistency, defined as a number of inliers $n$, has an exponential influence on the cosine distance ($d_L \leq 1$). If the amount of individuals in the database is large, re-identification can be made more efficient by using the aggregated Fisher vector for quick database searches, and using the geometric similarity only as the verification step.

\section{Experiments}
\label{sec:experiments}

The proposed method was evaluated with two different datasets: the SealID dataset~\cite{sealid} consisting of Saimaa ringed seals and the Whale shark dataset provided by Wild Me~\cite{holmberg2009estimating, sharkbook}. Saimaa ringed seal patterns consist of local arrangements of ring-like shapes. The regions enabling the re-identification often constitute a rather small portion of the whole pattern, making it important to have representative image features for local appearance. Whale shark patterns, on the other hand, consist of small spots with similar appearance,  making it more important to be able to quantify the geometric arrangement of the spots.

\subsection{Datasets}

\subsubsection{SealID}

\begin{figure}[!t]
    \centering
    \includegraphics[width=\linewidth]{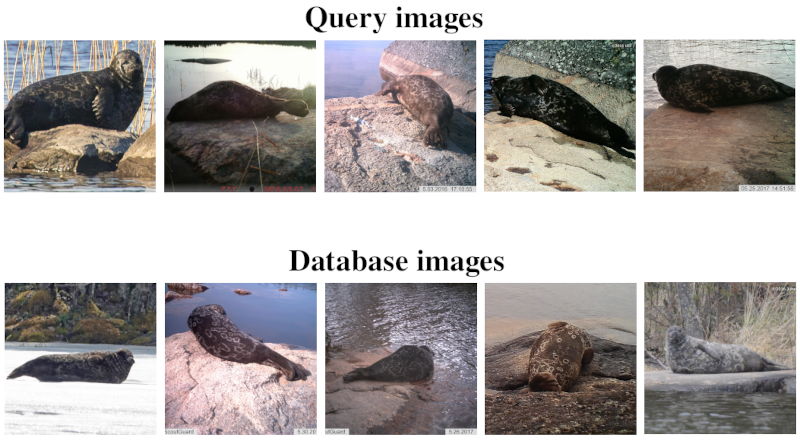}
    \caption{Sample images from both query and database images of the SealID dataset.}
    \label{fig:sealid}
\end{figure}

\begin{figure}[ht]
	\centering
	\minipage{0.5\linewidth}
	\includegraphics[width=\linewidth]{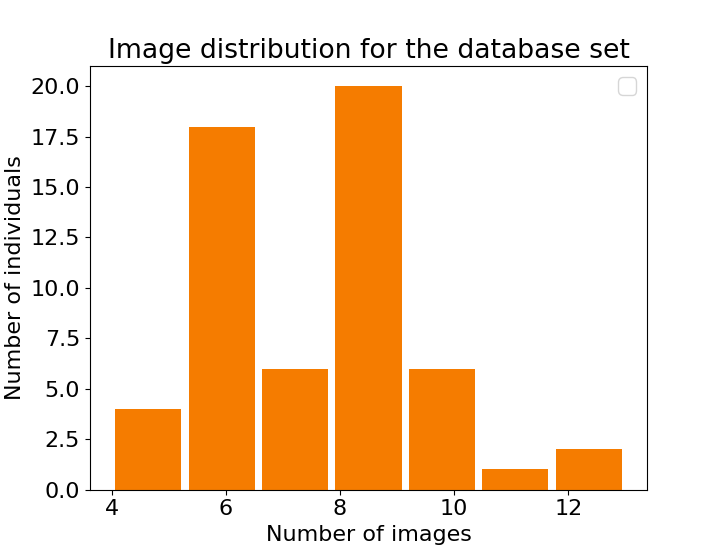}
	\endminipage\hfill
	\minipage{0.5\linewidth}
	\includegraphics[width=\linewidth]{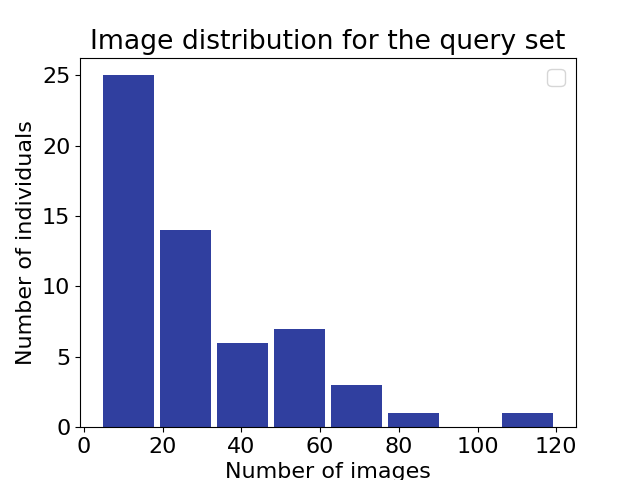}
	\endminipage\hfill
	\caption{Image distributions for the database and query sets. For example, in the query dataset (right) 25 individuals (y-axis) had less than 20 images (x-axis).~\cite{sealid}}
	\label{fig:seal_dist}
\end{figure}

The Saimaa Ringed Seal re-identification dataset (SealID) from~\cite{sealid} is used for the experiments. The dataset consists of 2080 images of 57 known Saimaa ringed seal individuals. The dataset is divided into two subsets: the database and the query. The database subset ($N=430$) consists of high-quality and unique images that are enough to cover the full body pattern of each individual seal in the dataset. The database has been constructed by prioritizing the images with the best quality. All training and preparations required for NORPPA are done using only the images from the database subset. The query subset ($N=1650$) contains the remaining images of the same seal individuals and these images are used in re-identification experiments. Sample images from both subset are shown in Fig.~\ref{fig:sealid}. The image distribution is shown in Fig.~\ref{fig:seal_dist}

\subsubsection{Whale shark dataset}

\begin{figure}[!t]
    \centering
    \includegraphics[width=\linewidth]{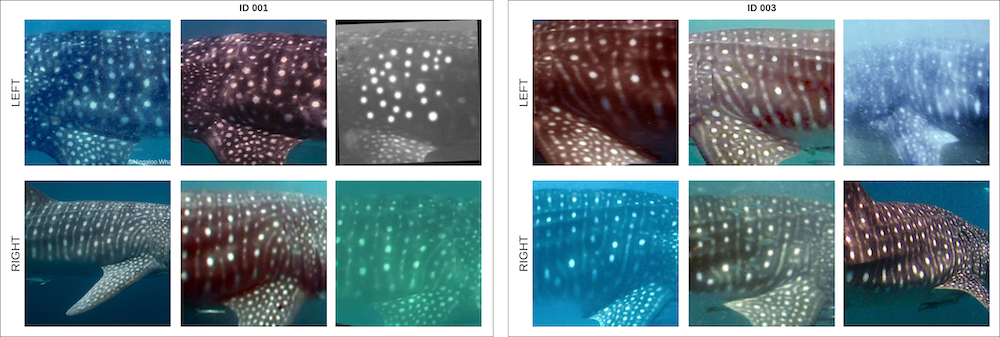}
    \caption{Sample images from the Whaleshark dataset.}
    \label{fig:whaleshark_examples}
\end{figure}

\begin{figure}[!t]
    \centering
    \includegraphics[width=\linewidth]{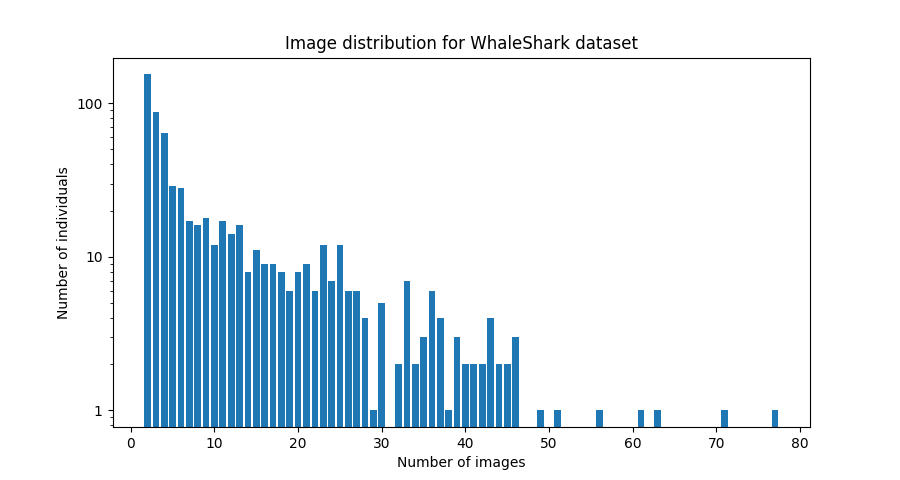}
    \caption{Image distribution for the Whaleshark dataset}
    \label{fig:whaleshark_dist}
\end{figure}

In the experiments, we utilized an extended version of the whale shark identification dataset provided by Wild~Me~\cite{holmberg2009estimating, sharkbook}. The original dataset includes images and corresponding labels in the Microsoft COCO format. Fig.~\ref{fig:whaleshark_examples} showcases example images from this dataset. Each image in the dataset is accompanied by a bounding box delineating the torso portion of the whale shark's body, an individual identification tag, and the viewpoint of the animal (right or left). The dataset comprises a total of 5409 annotated sightings, specifically pertaining to 235 distinct whale shark viewpoints. The image distribution is shown in Fig.~\ref{fig:whaleshark_dist}

%\subsection{Evaluation criteria}

\subsection{Segmentation and pattern extraction}

\begin{figure}[htp]
	\centering
	\subfloat[][]
	{
		\includegraphics[width=0.49\linewidth]{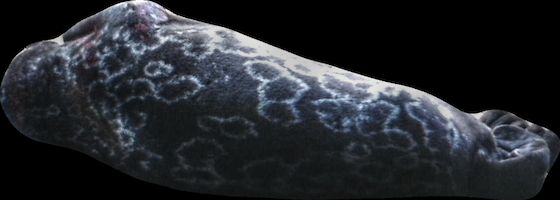}
		\label{subfig:pattern_original_seal}
	}
 	\subfloat[][]
	{
		\includegraphics[width=0.49\linewidth]{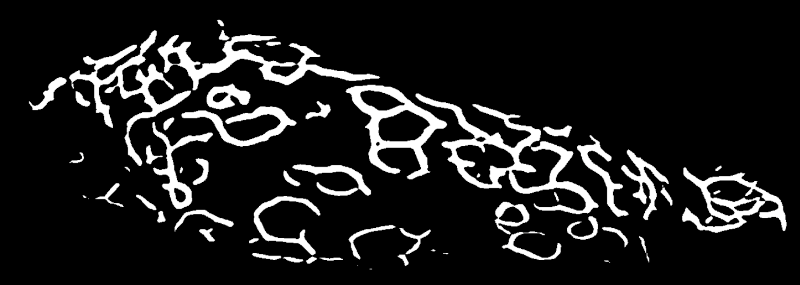}
		\label{subfig:pattern_seal}
	}\\
        \subfloat[][]
	{
		\includegraphics[width=0.32\linewidth]{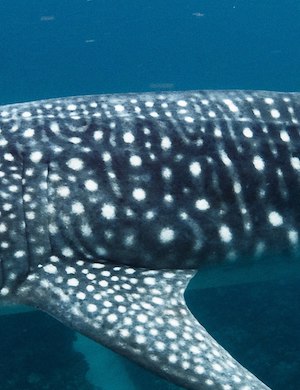}
		\label{subfig:pattern_norppa_original}
	}
	\subfloat[][]
	{
		\includegraphics[width=0.32\linewidth]{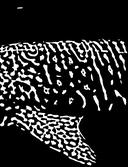}
		\label{subfig:pattern_norppa}
	}
        \subfloat[][]
	{
		\includegraphics[width=0.32\linewidth]{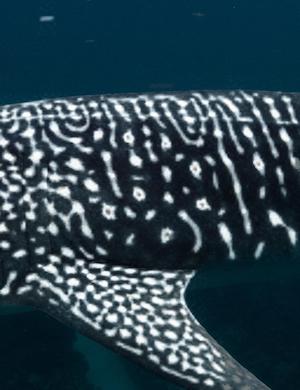}
		\label{subfig:mask_norppa}
	}\\
        \subfloat[][]
	{
		\includegraphics[width=0.32\linewidth]{figures/compressed/orig.jpeg}
		\label{subfig:pattern_original}
	}
	\subfloat[][]
	{
		\includegraphics[width=0.32\linewidth]{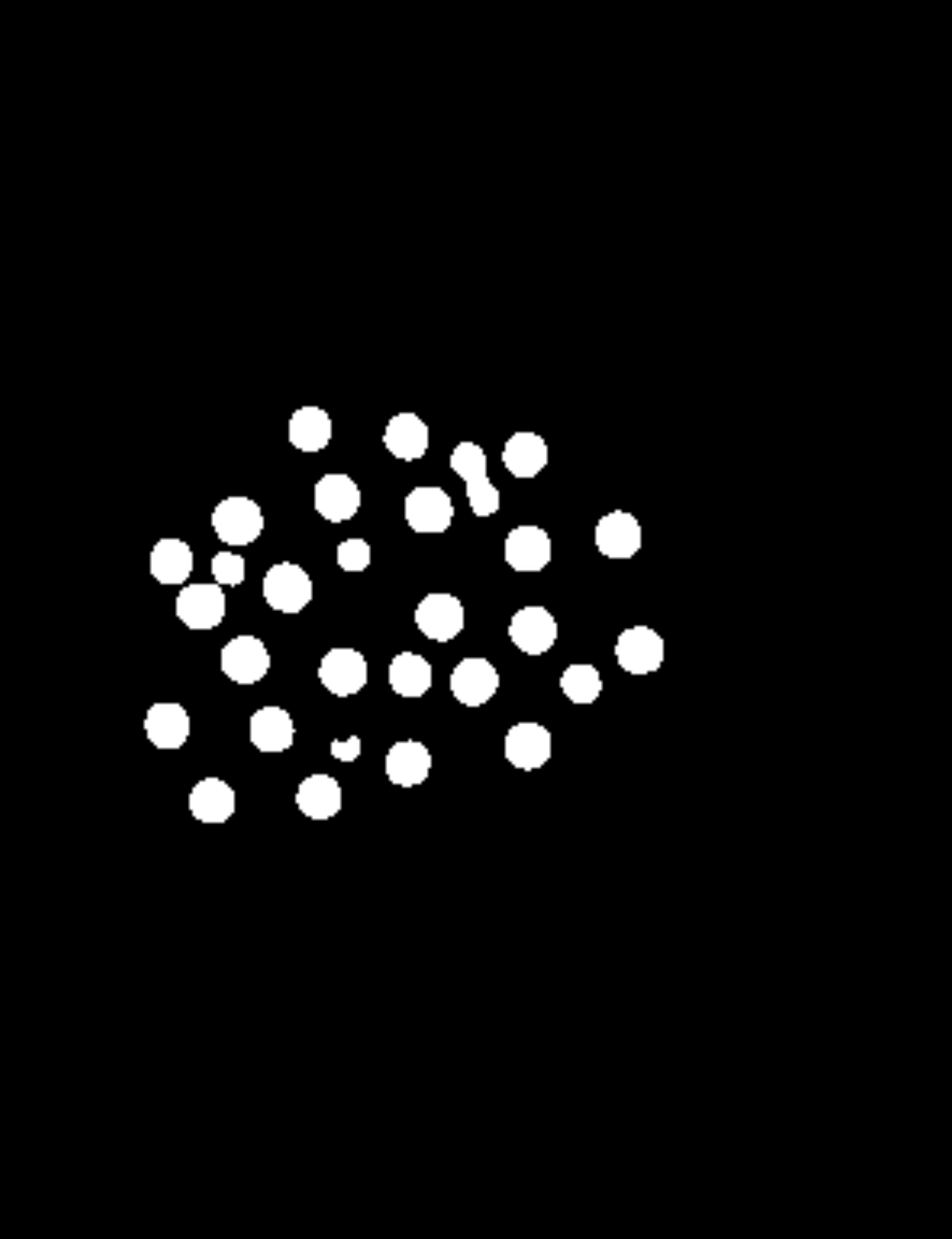}
		\label{subfig:pattern}
	}
        \subfloat[][]
	{
		\includegraphics[width=0.32\linewidth]{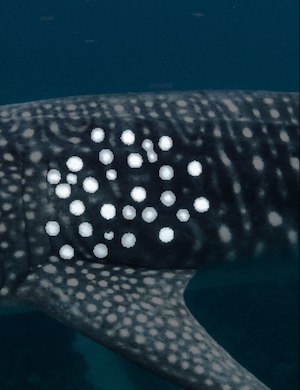}
		\label{subfig:mask}
	}
	\caption[]{Visualisation of the pattern extraction step for Saimaa ringed seals: (\subref{subfig:pattern_original_seal})-(\subref{subfig:pattern_seal}), whale sharks using the segmentation model trained on ringed seals: (\subref{subfig:pattern_original_seal})-(\subref{subfig:pattern_seal}), and whale sharks using the model from~\cite{kholiavchenko2022comprehensive}: (\subref{subfig:pattern_original})-(\subref{subfig:mask}).}
	\label{fig:pattern_ext}
\end{figure}

Saimaa ringed seals were segmented using Mask R-CNN with ResNet-101 backbone combined with Feature Pyramid Network trained on ringed seals. For more information about the segmentation method, see~\cite{ladoga}. 
For whale sharks, the segmentation step was omitted and the bounding box annotations were used. The evaluation, therefore, focused only on the re-identification. Pattern extraction for both datasets was performed using the U-Net-based model which was pretrained on Saimaa ringed seals. Since both patterns consist of white or light gray patterns on a dark background the same model was found to be reasonably accurate also on whale sharks as it can be seen from Fig.~\ref{fig:pattern_ext}. Furthermore, a U-Net-based spot segmentation method from~\cite{kholiavchenko2022comprehensive} specifically developed for whale sharks was tested (see Fig.~\ref{fig:pattern_ext}(f)-(h)).

\subsection{Saimaa ringed seal re-identification}

\begin{figure}[!ht]
% \begin{tikzpicture}
%     \begin{axis} [
%     width=.5\textwidth,
%     xlabel=$\text{top-}k$,
%     ylabel=$\text{accuracy (\%)}$,
%     xtick distance = 2,
%     ytick distance = 4,
%     ymin = 60, ymax = 95,
%     xmin = 0, xmax = 21,
%     grid = both,
%     %minor tick num = 1,
%     major grid style = {lightgray!25},
%     %minor grid style = {lightgray!50},
%     legend pos = south east,
%     %legend style={at={(.96,0.35)}, anchor=east, nodes={scale=.8,}, }
%     legend style={nodes={scale=.8,}, },
%     cycle list name=exotic,
%     line width=.4pt,
%     mark size=2.5pt,
%     mark repeat=2,
%     ]

%     \addplot file{data/norppa.dat};
%     % \addlegendentry{NORPPA~\cite{seal_cnn}: \(d_C = d_L\)}
%     \addlegendentry{NORPPA~[24]: \(d_C = d_L\)}

%     \addplot file{data/est_-n_0100.dat};
%     \addlegendentry{Only geometric similarity: \(d_C = -n\)}

%     \addplot file{data/est_d(1-w)_5_0100.dat};
%     \addlegendentry{Proposed: \(d_C = d_L (1 - \omega)^2\)}

%     \addplot file{data/est_d_n_0100.dat};
%     \addlegendentry{Proposed: \(d_C = d_L ^ n\)}

%     \addplot coordinates{(1, 61.9)(2, 62.75)(3, 63.6)(4, 64)(5, 64.4)};
%     % \addlegendentry{HotSpotter~\cite{crall2013hotspotter}}
%     \addlegendentry{HotSpotter~[4]}
    
%     \end{axis}
% \end{tikzpicture}
\includegraphics[clip, trim=0cm 23.1cm 12.5cm 0cm, width=1.00\linewidth]{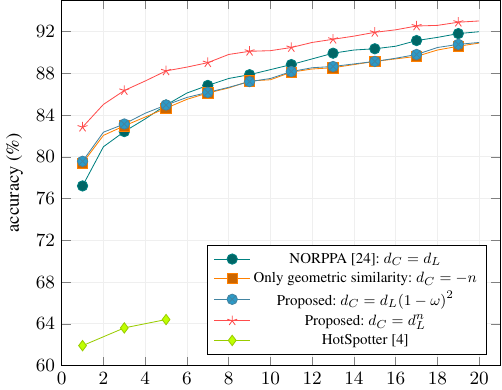}
\caption{Top-$k$ re-identification results relative to $k$ for SealID. For HotSpotter, the results were only analyzed through the top-5 accuracy.}
\label{fig:graph_seal}
\end{figure}

\begin{figure}[ht]
    \centering
    \begin{tabu}{c c c}
        \includegraphics[width=.269\linewidth]{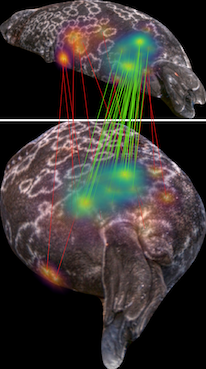} &
        \includegraphics[width=.362\linewidth]{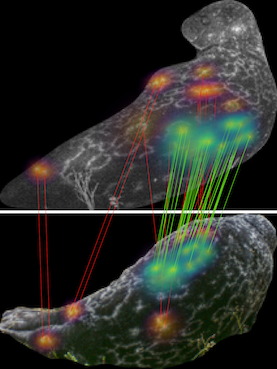} &
        \includegraphics[width=.285\linewidth]{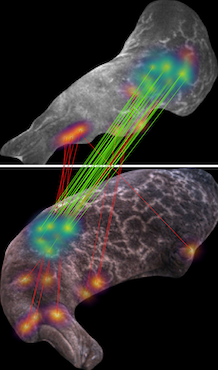} \\
        %(a) & (b) & (c) \\
        & &
    \end{tabu}
    
    \begin{tabular}{c c}
         \includegraphics[width=.462\linewidth]{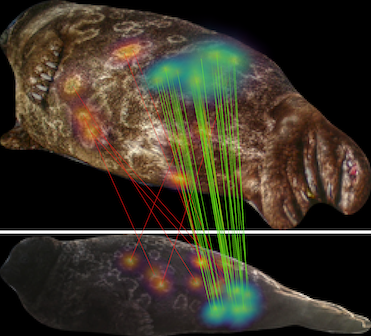} &
         \includegraphics[width=.482\linewidth]{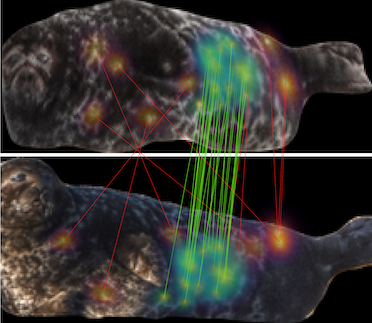} \\
         %(d) & (e)
    \end{tabular}

    \caption{Five examples of the successful Saimaa ringed seal re-identification by combination \(d_C = d_L ^ n\). The query image is at the top and the matched image from the database is at the bottom. Detected inliers are labeled in green and outliers are labeled in red.}
    %\caption{Five examples of top-1 incorrect-to-correct results by $\text{E}_{0.050}=d^n$. The query image is at the top and the corresponding image from the database is at the bottom. Inliers classified by~\gls{gva} are in green color and outliers are in red color.}
    \label{fig:inc_cor}
\end{figure}

\begin{figure}[ht]
    \centering
    \begin{tabu}{c c c}
        \includegraphics[width=.33\linewidth]{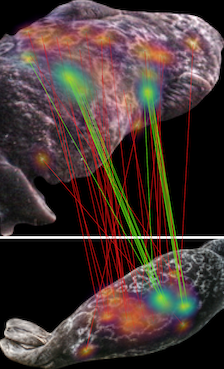} &
        \includegraphics[width=.287\linewidth]{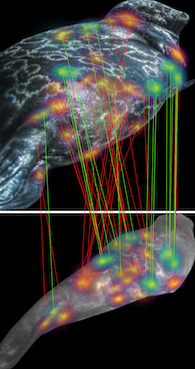} &
        \includegraphics[width=.266\linewidth]{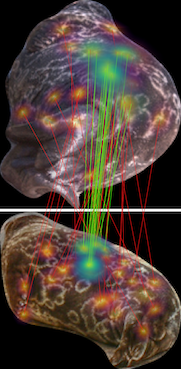} \\
        %(a) & (b) & (c) \\
    \end{tabu}
    \caption{Three examples of the unsuccessful re-identification by combination \(d_C = d_L ^ n\). The query image is at the top and the matched image from the database is at the bottom. Detected inliers are labeled in green and outliers are labeled in red.}
    %\caption{Three examples of correct-to-incorrect top-1 matches by $\text{E}_{0.050}=d^n$. The query image is at the top and the corresponding image from the database is at the bottom. Inliers classified by~\gls{gva} are in green color and outliers are in red color.}
    \label{fig:cor_inc}
\end{figure}

%In a total of 12 variations of the proposed method were tested on SealId dataset were tested. 
%In addition, re-identification results only with~\gls{norppa} were acquired in comparison.~\gls{gva} 
The re-identification results for the SealID dataset are presented in Table~\ref{tab:seal_results} and Fig.~\ref{fig:graph_seal}. NORPPA~\cite{seal_cnn} corresponds to using only similarity of aggregated local pattern appearances, that is $d_C = d_L$. When only geometric similarity is used the distance metric is defined by the number of inliers, that is $d_C = -n$. The RANSAC inlier threshold was set to 0.1 allowing a maximum displacement corresponding to 5\% of the size of the pattern image. A relatively large threshold value was used to take into account the large pose variation between images. 
The proposed combined pattern similarity measure outperformed the competing methods in the top-1 accuracy. The exponential combination rule provided the best re-identification accuracy. The re-identification results are illustrated in Fig.~\ref{fig:inc_cor}. The green lines correspond to the inliers found during the computation of geometric similarity. The red lines correspond to ouliers. In some rare cases (16 out of 1650 query images) the combined similarity failed to re-identify the individual even though NORPPA identified the individual correctly. This means that the RANSAC method found a notable amount of inliers despite the point correspondences being incorrect. Few of these cases are illustrated in Fig.~\ref{fig:cor_inc}.

\begin{table}[!h]
\processtable{Re-identification results on the SealID dataset.\label{tab:seal_results}}
{\begin{tabular*}{20pc}{@{\extracolsep{\fill}}llll@{}}\toprule
Method  & top-1  & top-3 & top-5 \\
\midrule
HotSpotter~\cite{crall2013hotspotter} &      61.9\%  & 63.6\%  &  64.4\%  \\
NORPPA~\cite{seal_cnn}: $d_C = d_L$    & 77.2\% & 82.4\% & 85.0\% \\ 
Only geometric similarity: $d_C = -n$ & 79.4\% & 83.0\% & 84.7\% \\
Proposed method: $d_C = d_L (1 - \omega)^2$ & 79.6\% & 83.2\% & 85.0\% \\
Proposed method: $d_C = d_L ^ n$ & \textbf{83.4\%} & \textbf{86.1\%} & \textbf{87.6\%} \\ 
\botrule
\end{tabular*}}{}
\end{table}

\subsection{Whale shark re-identification}

\begin{figure}[!ht]
% \begin{tikzpicture}
%     \begin{axis} [
%     width=.5\textwidth,
%     xlabel=$\text{top-}k$,
%     ylabel=$\text{accuracy (\%)}$,
%     xtick distance = 2,
%     ytick distance = 4,
%     ymin = 50, ymax = 85,
%     xmin = 0, xmax = 21,
%     grid = both,
%     %minor tick num = 1,
%     major grid style = {lightgray!25},
%     %minor grid style = {lightgray!50},
%     legend pos = south east,
%     legend style={nodes={scale=.8,}, },
%     cycle list name=exotic,
%     line width=.4pt,
%     mark size=2.5pt,
%     mark repeat=2,
%     ]

%     \addplot file{data/ws_norppa.dat};
%     % \addlegendentry{NORPPA~\cite{seal_cnn}: \(d_C = d_L\)}
%     \addlegendentry{NORPPA~[24]: \(d_C = d_L\)}

%     \addplot file{data/ws_geo_0050.dat};
%     \addlegendentry{Only geometric similarity: \(d_C = -n\)}
    
%     \addplot file{data/ws_pol_0050_5000.dat};
%     \addlegendentry{Proposed: \(d_C = d_L (1 - \omega)^2\)}

%     \addplot file{data/ws_exp_0020.dat};
%     \addlegendentry{Proposed: \(d_C = d_L ^ n\)}

%     \addplot coordinates{(1, 52)(2, 52.5)(3, 53)(4, 53)(5, 53)};
%     % \addlegendentry{HotSpotter~\cite{crall2013hotspotter}}
%     \addlegendentry{HotSpotter~[4]}
    
%     \end{axis}
% \end{tikzpicture}
\includegraphics[clip, trim=0cm 23.1cm 12.5cm 0cm, width=1.00\linewidth]{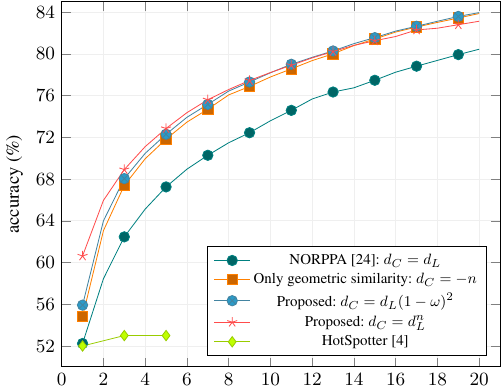}
\caption{Top-$k$ re-identification accuracy relative to $k$ for the Whale shark dataset.}
\label{fig:graph_whale}
\end{figure}

\begin{figure}[ht]
    \centering
    \begin{tabu}{c c c}
        \includegraphics[width=.33\linewidth]{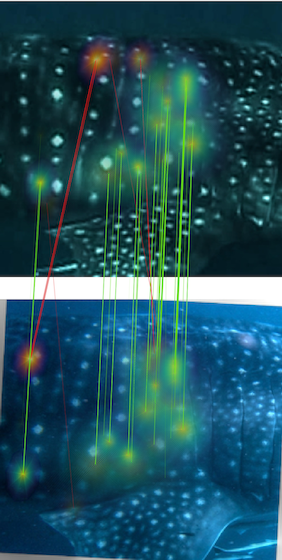} &
        \includegraphics[width=.285\linewidth]{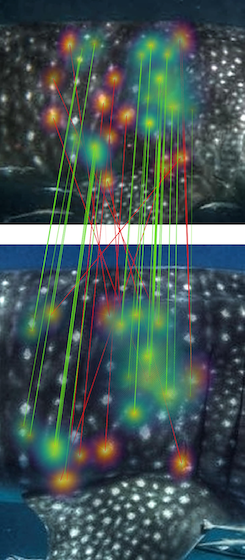} &
        \includegraphics[width=.26\linewidth]{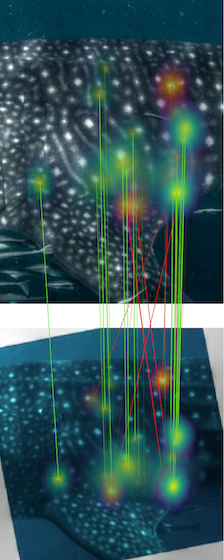} \\
        %(a) & (b) & (c) \\
    \end{tabu}
    \caption{Examples of the successful whale shark re-identifications using combination \(d_C = d_L ^ n\).}
    %\caption{Three examples of correct-to-incorrect top-1 matches by $\text{E}_{0.050}=d^n$. The query image is at the top and the corresponding image from the database is at the bottom. Inliers classified by~\gls{gva} are in green color and outliers are in red color.}
    \label{fig:shark_results}
\end{figure}

The whale shark dataset was not divided into the database and query subsets. Instead, whale shark re-identification was performed using a leave-one-out strategy. The RANSAC inlier threshold was set to 0.05. A smaller threshold value was used than for Saimaa ringed seals due to the smaller variation in pose and the need to find finer dissimilarities on spot arrangements. A general rule of thumb when adjusting the threshold for new species is that it can be stricter for cases where the transformation between the patterns is more linear or suitable to be described with a homography, which is the case for the whalesharks. The results are shown in Table~\ref{tab:whaleshark_results} and Fig.~\ref{fig:graph_whale}. The successful re-identification results are visualized in Fig.~\ref{fig:shark_results}. As it can be seen the proposed method outperforms both HotSpotter and NORPPA. While the accuracies are notably lower than for Saimaa ringed seals, the results can be considered promising since the same pattern extraction and feature embedding models as for the ringed seals were used without retraining or fine-tuning. By simply replacing the pattern extraction model with the whale shark-specific spot segmentation method~\cite{kholiavchenko2022comprehensive}, the top-1 accuracy increased to 73\%. Fully omitting the pattern extraction step lowered the top-1 accuracy to 59\% showing that even suboptimal pattern extraction method still improves the accuracy.

\begin{table}[!h]
\processtable{Re-identification results on the whale shark dataset.\label{tab:whaleshark_results}}
{\begin{tabular*}{20pc}{@{\extracolsep{\fill}}llll@{}}\toprule
Method  & top-1  & top-3 & top-5 \\
\midrule
HotSpotter~\cite{crall2013hotspotter} &  52\%     & 53\% &  53\% \\
NORPPA~\cite{seal_cnn}: $d_C = d_L$    & 53\% & 62\% & 69\% \\ 
Only geometric similarity: $d_C = -n$ & 55\% & 67\% & 72\% \\
Proposed: $d_C = d_L (1 - \omega)^2$ & 56\% & 68\% & 72\% \\
Proposed: $d_C = d_L ^ n$ & \textbf{61\%} & \textbf{69\%} & \textbf{73\%} \\ 
\botrule
\end{tabular*}}{}
\end{table}

% \begin{figure}[!t]
% \centering{\includegraphics{Sample_1.eps}}
% \caption{Sample graph with blue (dotted), green (solid) and red (dashed) lines
% \figfooter{a}{Subfigure 1}
% \figfooter{b}{Subfigure 2}\label{fig:one}}
% \end{figure}
% \begin{figure*}[!b]
% \centering{\includegraphics{Sample_2.eps}}
% \caption{Sample graph spaning two columns\label{fig:two}}
% \end{figure*}
% The preferred format is encapsulated postscript (.eps) for line figures and .tif for halftone figures with a minimum resolution of 300 dpi (dots per inch).

% \begin{table}[!b]
% \processtable{Example table\label{tab1}}
% {\begin{tabular*}{20pc}{@{\extracolsep{\fill}}lll@{}}\toprule
% Column  &Column  & Column heading \\
% heading  &heading two &  three \\
% \midrule
% Row 1a  &Row 1b  &Row 1c \\
% Row 2a  &Row 2b  &Row 2c \\
% Row 3a  &Row 3b  & Row 3c \\
% Row 4a  &Row 4b  &Row 4c \\
% Row 5a  &Row 5b  &Row 5c \\
% Row 6a  & Row 6b  & Row 6c \\
% \botrule
% \end{tabular*}}{}
% \end{table}

\section{Conclusion}\label{sec:conclusion}

In this paper, the re-identification of animal individuals using unique fur and skin patterns has been considered. The proposed method combines both similarity of local visual appearances of the pattern aggregated over the full image, as well as the global geometric consistency of the pattern similarities. This provides a versatile pattern similarity metric that can be used for re-identification on a wide variety of patterned animals. We demonstrated the method on two species with notably different types of patterns: Saimaa ringed seals and whale sharks. Promising results were obtained with the combined similarity providing higher accuracy than for the individual pattern similarity metrics. One notable benefit of the proposed method is that species-specific training is not necessarily needed. To demonstrate this, the same pre-trained models were used on both species for pattern extraction and pattern feature embedding making the proposed method species-agnostic without additional training. At same time, the models are trainable making it possible to fine-tune the method for a specific species. This has potential to further increase the re-identification accuracy.

\section{Acknowledgments}\label{sec:acknowledgments}

The authors would like to thank Vincent Biard, Piia Mutka, Marja Niemi, and Mervi Kunnasranta from the Department of Environmental and Biological Sciences at the University of Eastern Finland (UEF) for providing the data of Saimaa ringed seals and their expert knowledge of identifying each individual. The authors would like to thank Maksim Kholiavchenko from the Department of Computer Science, Rensselaer Polytechnic Institute, for providing additional insight into their method.

% section example_references (end)

% \vfill\pagebreak

% \section{Appendices}\label{sec14}

% Additional material, e.g. mathematical derivations, tables and figures
% larger than half a page that may interrupt the flow of your paper's argument
% should form a separate Appendix section (see Table~\ref{tab2}). Do not, however, use
% appendices to lengthen your article unnecessarily as this section is
% included in the word count. If the material can be found in another work,
% cite this work rather than reproduce~it.
% The appendix section should be in double column format, and come after the references.

%%%%%%%%%% end main matter %%%%%%%%%%

%%%%%%%%%%%%%% bibliography %%%%%%%%%%
\bibliographystyle{plainnat}
% \bibliography{references}

%%%%%%%%%% end bibliography %%%%%%%%%%

\end{document}